\documentclass[letterpaper, 10 pt, conference]{ieeeconf}  

\usepackage{booktabs}
\usepackage{multirow}
\usepackage{graphics} 
\usepackage{epsfig} 
\usepackage{mathptmx} 
\usepackage{times} 
\usepackage{amsmath,amssymb,amsfonts}
\usepackage{algorithm2e}
\usepackage{cite}
\usepackage{subfigure}
\usepackage{multirow}
\def\statespace {{\cal S}}
\def\mdp {{\cal M}}
\def\actionspace {{\cal A}}
\def\probs {{\cal P}}
\def\transitionmodel {{\cal T}}

\IEEEoverridecommandlockouts                              

\newcommand{\real}{\mathbb{R}}

\title{\LARGE \bf
\textmd{DEUX}: Active Exploration for Learning Unsupervised Depth Perception
}

\author{Marvin Chanc\'an, Alex Wong and Ian Abraham
\thanks{Yale University, CT 06510, USA. {\tt\small marvin.chancan@yale.edu}}
}

\begin{document}

\maketitle
\thispagestyle{empty}
\pagestyle{empty}

\begin{abstract}
    Depth perception models are typically trained on non-interactive datasets with predefined camera trajectories. 
    However, this often introduces systematic biases into the learning process correlated to specific camera paths chosen during data acquisition. In this paper, we investigate the role of how data is collected for learning depth completion, from a robot navigation perspective, by leveraging 3D interactive environments. First, we evaluate four depth completion models trained on data collected using conventional navigation techniques. Our key insight is that existing exploration paradigms do not necessarily provide task-specific data points to achieve competent unsupervised depth completion learning. We then find that data collected with respect to photometric reconstruction has a direct positive influence on model performance. As a result, we develop an active, task-informed, depth uncertainty-based motion planning approach for learning depth completion, which we call \textmd{DE}pth \textmd{U}ncertainty-guided e\textmd{X}ploration (\textmd{DEUX}). Training with data collected by our approach improves depth completion by an average greater than 18\% across four depth completion models compared to existing exploration methods on the MP3D test set. We show that our approach further improves zero-shot generalization, while offering new insights into integrating robot learning-based depth estimation.
\end{abstract}

\section{INTRODUCTION} \label{sec:intro}

\begin{figure}[!t]
    \includegraphics[width=\columnwidth]{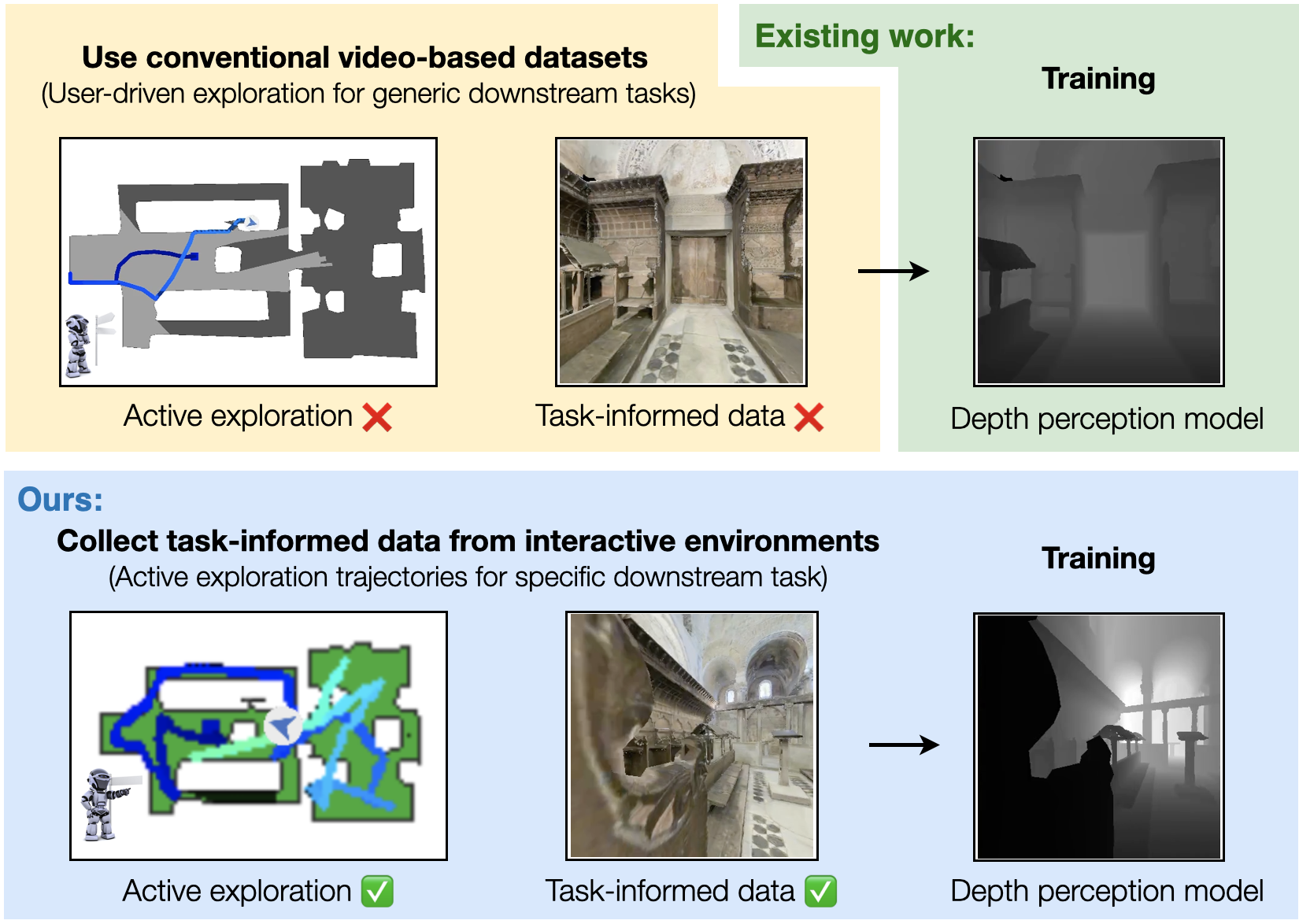}
    \caption{\textbf{The depth perception problem}. Existing approaches typically use publicly available video-based datasets for training depth perception models (top). Our active exploratory approach for task-informed data collection (bottom) improves depth completion results yield by existing exploration paradigms, and zero-shot generalization to other datasets.}
    \label{fig1}
    \vspace{-5mm}
\end{figure}

    Depth estimation supports a wide range of robotic and computer vision applications including autonomous navigation, augmented reality, and three-dimensional (3D) mapping, planning and reconstruction. 
    Recent advances in unsupervised learning for depth estimation from a single RGB camera and depth sensor (\textit{e.g.}, 3D LiDAR) have enabled the supervision signal to scale w.r.t. data size, \textit{i.e.}, autonomous data collection. 
    Despite these rapid advances, existing depth perception models have largely been trained on non-interactive data-sets, \textit{e.g.}, collected from non-autonomous, user-driven camera trajectories with predetermined routes, such as KITTI \cite{geiger2012we, uhrig2017sparsity} for outdoor scenarios, or VOID \cite{wong2020unsupervised} and NYUv2 \cite{silberman2012indoor} for indoors. 
    While their purpose was to facilitate learning for a variety of vision tasks, they introduce systematic errors on account of the lack of task-specific diversity that is often seen during real-world deployment (Fig. \ref{fig1}-top). 
    Even with the option of revisiting the data collection sites, determining what more training data to collect based on some test set  remains in question. While one can attempt to densely/uniformly collect data from the environment, as opposed to actively sampling key data points w.r.t. utility metrics, it is not scalable in real scenarios. 
    
    Motivated by these shortcomings, we investigate the influence of robotic exploration on the performance of unsupervised depth completion, the task of inferring dense depth from an image and a synchronized sparse point cloud, by (i) using interactive environments for task-specific data collection, and (ii) proposing an active exploration approach driven by model error modes, instead of relying on user-based or conventional navigation paradigms (Fig. \ref{fig1}-bottom). 
    

Robotic navigation, on the other hand, has a long history in both classic and learning-based techniques designed to enable high-level motion planning and localization tasks. Although these two lines of research have been largely disconnected, recent works have attempted to create a standard framework for benchmarking classical and end-to-end learning-based exploration methods by using complex interactive environments \cite{mishkin2019benchmarking, ramakrishnan2021exploration} or proposing active viewpoint sampling \cite{chen2022learning, cao2021tare}. In contrast to these works, which mainly study navigation tasks, here we focus on exploration for learning depth completion specifically in the context of robot navigation and task-driven data collection. We propose an active, depth uncertainty-guided exploration approach, which integrates 3D reconstruction errors, to allow task-driven exploratory planning and further improve the overall depth estimation performance. Extensive experimental results show that our approach can provide rich and diverse data that improves the state-of-the-art of unsupervised depth completion models. The main contributions of this paper are:



\begin{itemize}
    \item A benchmark on the use of classic and learning-based navigation methods for training supervised and unsupervised depth completion models.
    \item An active, task-informed exploration approach to guide robot motion planning, via photometric reprojection errors, for data collection to improve depth estimation. 
\end{itemize}

\section{RELATED WORK} \label{sec:related_work}

\subsection{Autonomous Robotic Exploration}

\textbf{Conventional Exploration}. The goal of autonomous exploration algorithms is to allow robotic platforms to navigate across regions of interest for specific downstream tasks within an environment (\textit{i.e.}, single or multi-goal localization \cite{chancan2022role}, and area coverage for 3D mapping/reconstruction \cite{schoenberger2018thesis}). The task of area coverage, where a robot visits every space of an environment, requires certain exploratory capabilities such as directed lawn-mowers~\cite{rekleitis2008efficient, rekleitis2005distributed, choset2001coverage, yu2019coverage}, random walks~\cite{pang2021effect, carpin2005motion, tan2022embodied}, information-based methods~\cite{amigoni2010information, jadidi2015mutual, nelson2015information, girdhar2014curiosity, lerch2023safety, dong2023time, wittemyer2023bi}, path-planning optimization via viewpoint sampling \cite{cao2021tare,yang2022far}, or semantic-based exploration \cite{li2022comparison,ramakrishnan2022poni, li2022learning}. Most of these exploration methods, however, are typically designed to ensure space coverage, semantic mapping, or point/object-goal search instead of specific downstream computer vision tasks such as depth perception. This limits autonomous exploration of important areas with equal proportion relative to an expected utility value of the downstream task. To this end, here we focus on studying the influence of robotic motion and exploratory approaches in the context of depth perception learning.


\textbf{Interactive Environments}. Recent work uses interactive environments for (i) benchmarking classic and learned navigation methods in single-goal localization tasks \cite{mishkin2019benchmarking, ramakrishnan2021exploration}, or (ii) learning active camera exploration for multi-goal localization tasks \cite{chen2022learning}. In \cite{mishkin2019benchmarking}, the authors demonstrate that classical and end-to-end learning-based exploration approaches are still far from human-level performance in single-goal navigation tasks (\textit{e.g}, in terms of Success weighted by Path Length (SPL), success rate, and pace). Also for single-goal navigation tasks, \cite{ramakrishnan2021exploration} provides extensive classic and learned navigation results while introducing new exploration paradigms based on reinforcement learning (\textit{e.g}, area coverage, novelty, curiosity, and semantic-based reconstruction). Furthermore, \cite{chen2022learning} proposed active learning for multi-goal navigation, however this approach exploits human experience to guide the active learning process.


For depth completion tasks, there is existing work that also uses 3D simulation environments \cite{jeon2022struct,senushkin2021decoder,kam2022costdcnet,tan2021mirror3d,wu2022toward}, but these do not explore the influence of robot motion and exploration into the learning process. A recent dataset for robot perception and navigation was proposed in \cite{wang2020tartanair}, which is collected from 3D environments. However, this work uses pre-defined camera trajectories, obtained by incremental mapping and trajectory sampling via RRT$^*$ \cite{noreen2016comparison}, specifically from collision-free navigation tasks only. Our focus in this work, instead, is exploring the influence of diverse navigation methods for data collection to improving depth completion performance. Our proposed framework also aims to supporting new research on the influence of robot exploration into the performance of any other perception modality algorithm.

\subsection{Depth Estimation}

Depth can be inferred from various sources, \textit{i.e.}, stereo \cite{chang2018pyramid,duggal2019deeppruner,poggi2020self,xu2020aanet}, multi-view \cite{chen2019point,gu2020cascade,wang2021patchmatchnet,yao2018mvsnet,yao2019recurrent}, and monocular \cite{fei2019geo,godard2019digging,poggi2022real,ranftl2021vision,watson2019self,wong2019bilateral} images, which can also be used in combination with sparse range \cite{hu2021penet,liu2022monitored,long2021radar,park2020non,singh2023depth,wong2021learning,wong2021adaptive,wong2021unsupervised,yang2019dense}. We focus on monocular depth completion, which aims to infer a 2.5D dense depth map from a single image and a synchronized sparse point cloud---supporting dense mapping in the exploration task. With the need to localize the agent, \textit{e.g.}, a robot during exploration, one typically employs a simultaneous localization and mapping (SLAM) \cite{davison2003real,engel2014lsd,mur2015orb} or a visual inertial odometry (VIO) \cite{jones2007inertial,jones2011visual,geneva2020openvins,mourikis2007multi} system, which tracks a sparse point cloud. While the point cloud is sufficient for localization, it is far too sparse for representing the structure of 3D environments. Hence, we choose depth completion to densify or complete the point cloud with guidance from a single image, which naturally integrates well with SLAM or VIO systems \cite{merrill2021robust,wong2020unsupervised}. This form of depth estimation supports inputs from any of the aforementioned streams without concerns for accumulating sufficient parallax (\textit{i.e.}, multi-view) obtaining scale (\textit{i.e.}, monocular) or requiring an additional camera (\textit{i.e.}, stereo) at test time. Our choice in depth completion methods belongs to the unsupervised learning paradigm, so that we do not assume access to ground truth depth for training, but only calibrated images and their associated point clouds that are measured by a minimal (optical, inertial) sensor setup.

\textbf{Unsupervised depth completion} training typically relies on supervision based in structure-from-motion, whether from stereo or monocular video. Methods trained with stereo  \cite{shivakumar2019dfusenet,yang2019dense} require rectified stereo pairs and predict disparity between the two frames. The supervision signal comes from reconstructing each frame from other other and ensuring left-right consistency between the reconstructions; depth is inversely proportion to disparity and can be obtained in closed form using the focal length and baseline between the stereo cameras. Similarly, methods that leverage monocular video \cite{ma2019self,liu2022monitored,wong2021learning,wong2021adaptive,wong2020unsupervised} minimize forward-backward reconstruction error from a subset of frames in a video to a given  temporally nearby reference frame of the same video. To this end, methods typically jointly optimize for the predicted depth and relative pose between the videos frames. Both stereo and monocular video training modes reconstructs the sparse point cloud as an additional loss term to ground estimates to metric scale. As 3D reconstruction from 2D image and sparse range measurements is an ill-posed problem, the training objective also includes a local smoothness regularizer. In this work, all of the depth completion models chosen rely on video-based training, which requires a single camera, and if available, inertial measurement unit (IMU); both are ubiquitous in most devices and suitable for deployment with SLAM and VIO systems.


\section{THE DEUX APPROACH} \label{sec:methods}



\subsection{Learning Unsupervised Depth Completion}

Given an RGB image $I : \Omega \subset \real^2 \mapsto \real^3_+$ and its sparse point cloud (projected onto the image plane) $z : \Omega_z \subset \Omega \mapsto \real_+$, we learn a function $h_\theta(I, z) : \Omega \mapsto \real_+$, parameterized by $\theta$, that maps the image and sparse depth into a dense depth map. To train $h_\theta(I, z)$, we assume access to temporally consecutive frames, \textit{i.e.}, a monocular video, at time $t$, $t-1$, and $t-2$ and the camera intrinsic calibration matrix $K \in \real^{3\times 3}$. We minimize the photometric reprojection error (Eqn. \ref{eqn:photometric_reprojection_loss}) between $I_{t}$ and its reconstructions $\hat{I}_{t, t-1}$ and $\hat{I}_{t, t-2}$ from $I_{t-1}$ and $I_{t-2}$, where each reconstruction $\hat{I}_{t, \tau}$ is obtained via


\begin{equation}
\label{eqn:image_reconstruction}
    \hat{I}_{t, \tau}(q, \hat{d}(x), p_{\tau, t}) = I_{\tau} \big( \pi  \ p_{\tau, t} \ K^{-1} \ \bar{q} \  \hat{d}(q) \big),
\end{equation}
where $\hat{d} := h_\theta(I, z)$ for ease of notation, $\tau \in \{t-1, t-2\}$ the time step, $\bar{q} = [q^\top 1]^\top$ the pixel location as a homogeneous coordinate, $p_{\tau, t} \in SE3$ is the relative camera motion from time $t$ to $\tau$ and $\pi$ the canonical perspective projection. Specifically, the photometric reprojection loss is comprised of color consistency and structural consistency terms. Color consistency penalizes the $L1$ difference between $I_{t}$ and $\hat{I}_{t, \tau}$: 
\begin{equation}
\label{eqn:color_consistency_loss}
  	\ell_{co}(I_{t}, \hat{I}_{t,\tau}) = \frac{1}{|\Omega|} \sum_{q \in \Omega} 
  	|\hat{I}_{\tau}(q)-I_{t}(q)|.
\end{equation}
Structural consistency measures the structural similarity between $I_{t}$ and $\hat{I}_{t, \tau}$ using $\texttt{SSIM}$ \cite{SSIM}. We subtract $\texttt{SSIM}$ score from 1 to penalize $\hat{I}_{t, \tau}$ for deviations from $I_{t}$: 
\begin{equation}
\label{eqn:structural_consistency_loss}
  	\ell_{st}(I_{t}, \hat{I}_{t,\tau}) = \frac{1}{|\Omega|} \sum_{q \in \Omega} 
  	\big(1 - \texttt{SSIM}(\hat{I}_{\tau}(q), I_t(q))\big).
\end{equation}

The photometric reprojection loss is their linear combination weighted by their respective $\lambda$ summed over all frames:
\begin{equation}
\label{eqn:photometric_reprojection_loss}
  	\ell_{ph} = \sum_{\tau \in \{t-1, t-2 \}} 
  	    \lambda_{co} \ell_{co}(I_{t}, \hat{I}_{t,\tau}) + \lambda_{st} \ell_{st}(I_{t}, \hat{I}_{t,\tau}).
\end{equation}
Because 3D reconstruction from 2D images is an ill-posed problem, the use of a regularizer to enforce local smoothness and connectivity over $\hat{d}$ is needed, following \cite{wong2021adaptive,wong2021unsupervised}:
\begin{equation}
\label{eqn:local_smoothness_loss}
    \ell_{sm} = \frac{1}{|\Omega|}
        \sum_{q \in \Omega} 
            e^{-|\nabla I_{t}(q)|}| \nabla \hat{d}(q)|,
\end{equation}
where the gradient of $\hat{d}$ is weighted by the image gradient to allow for depth discontinuities across object boundaries.

Minimizing the reprojection error will reconstruct the scene structure up to an unknown scale \cite{wong2021unsupervised}. Predictions are grounded to metric scale by minimizing the $L1$ difference between $\hat{d}$ and $\Omega_z$ (over its domain), as follows:
\begin{equation}
\label{eqn:sparse_loss}
    \ell_{sz} = \frac{1}{|\Omega_z|}
        \sum_{q \in \Omega_z} 
            |\hat{d}(q) - z(q)|.
\end{equation}

The unsupervised depth estimation loss, therefore, reads:
\begin{equation}
\label{eqn:unsupervised_depth_loss}
  	\ell_{d} = \ell_{ph} + \lambda_{sz} \ell_{sz} + \lambda_{sm} \ell_{sm}.
\end{equation}

\subsection{Uncertainty Measures in Depth Estimation}

Measuring uncertainty in depth estimation tasks in typically achieved via image reconstruction \cite{hu2012quantitative,yang2020d3vo,wong2021adaptive}. But rather than using the matching cost as a proxy for uncertainty measure \cite{hu2012quantitative} or to guide depth completion learning \cite{wong2021adaptive}, we use it to provide us with specific scene locations that are likely to contain high depth uncertainty. For instance, due to inconsistent robot motions or inherently challenging scenes, more informative data can be collected from robot exploration around these areas to address potential failure modes of the model. We assume a set of pretrained weights obtained from minimizing Eqn. \ref{eqn:unsupervised_depth_loss} in Sect. \ref{sec:methods}. To determine the error modes of a model, we compute the photometric reprojection errors between $I_{t}$ and its reconstructions $\hat{I}_{t, t-1}$ and $\hat{I}_{t, t-2}$ from $I_{t-1}$ and $I_{t-2}$ using Eq. \ref{eqn:image_reconstruction}. We then compute the mean of these errors over all pixels in the image to obtain a scalar residual value $\delta_{\tau}(q)$ for each $I_{\tau}(q)$ as follows:

\RestyleAlgo{ruled}
\SetKwComment{Comment}{/*}{*/}
\begin{algorithm}[t!]
\caption{Depth Uncertainty-guided Nav. Policy}\label{alg:depth_exploration_algo}
\KwData{map $\mathfrak{M}$, time budget $T_{exp}$, residual pose $p_{\delta}$}
\While{$T_{exp}$ not reached}{
    $p_f$ $=$ SampleFrontiers($\mathfrak{M}$)\;
    $p^{tgt} = $ SampleDepthGuidedTarget($p_f$, $p_\delta$)\;
    \While{not reached $p_{\delta}$}{
        $\mathfrak{\hat{M}} = $ ProcessMap($\mathfrak{M}$)\;
        $Path_{tgt}$, $\Delta_{next}$ $=$ AStarPlanner($\mathfrak{\hat{M}},$ $p^{tgt}$)\;
        \eIf{Reached $p^{tgt}$}{
            break\;
            }{
            \eIf{$Path_{tgt}$ is not None}{
                $p^{next}=$ Path$[\Delta_{next}]$\;
                action $=$ get\_action($p^{next}$)\;
            }{
                action $=$ random()\Comment*[r]{unstuck}
            }
            }
    }
}
\end{algorithm}

\begin{equation}
\label{eqn:depth_uncertainty}
    \delta_{\tau}(q) = 1- e^{-\frac{1}{|\Omega|} \sum_{q \in \Omega} |\hat{I}_{\tau}(q)-I_{t}(q)|}
\end{equation}
we use $\delta_{\tau}(q) \in \real_+$ with $L1$ penalty as a proof-of-concept, but it can be replaced by other metrics, \textit{i.e.}, \texttt{SSIM} \cite{wong2021adaptive}.

\subsection{Robot Exploration}

Our goal is to obtain an exploration policy to perform robot navigation tasks. We use a Markov Decision Process $\mdp$ with discrete states $\mathbf{s}_t \in \statespace$ and actions $\mathbf{a}_t \in \actionspace$ spaces, with a transition operator $\transitionmodel: \statespace \times \actionspace \to \statespace$ to model our navigation task as a finite-horizon $T$ problem. Our navigation policy will maximize the objective function given by

\begin{equation}
\label{objective}
J = \mathbb{E}_{\tau \sim \pi_{\tau}} \left[\sum_{t=1}^T r(\tau)\right]
\end{equation}

where $\pi: \statespace \to \probs(\actionspace)$ is the navigation policy we want to design (reported in Alg. 1), and $r: \statespace \times \actionspace \to \mathbb{R}$ is a reward function we want to maximize (given by the negative of the Eq. \ref{eqn:depth_uncertainty}). We define $\pi$ as an algorithm based on our residual values $\delta_{\tau}(q)$, as described in the next section. 

\subsection{Depth Uncertainty-guided Exploration (\textmd{DEUX})}

We consider a robot equipped with a depth estimation model, which is instantiated to navigate an environment, while computing $\delta_{\tau}(q)$ for every time step. We note that this exploration stage is limited by a maximum number of time steps $T_{exp}$, and that the agent is spawned at an initial pose $p_0 \in SE3$, pre-defined by the 3D simulator configuration. Given the dense map outputs of the depth perception model $\hat{d}_{\tau}(q)\in \real^{h\times w}_+$, we follow \cite{yamauchi1997frontier, ramakrishnan2021exploration} to build a 2D top-down occupancy map of the environment $\mathfrak{M}\in \real^{h\times w}$, which indicates whether a certain $(x, y)$ location is navigable or occupied. This egocentric occupancy map $\mathfrak{M}$ is used to compute the frontiers between free and occupied space, which along with $\delta_{\tau}(q)$ are used to generate the next depth-informed target locations. A target is chosen one at a time by prioritizing regions with high uncertainty. When a target is selected, the A-Star planner algorithm \cite{hart1968formal} is used to process the map $\mathfrak{M}$ and generate the shortest path from the current position of the robot. Once the robot reaches the desired location, the next target is sampled along with the path to navigate there, following Alg. \ref{alg:depth_exploration_algo}.

\section{EXPERIMENTS} \label{sec:experiments}

\subsection{Experimental Setup}

\textbf{Interactive Environments}. We leverage the Habitat-Sim simulator \cite{savva2019habitat} and its Habitat-Lab API as our experimental platform for embodied robot exploration and data collection. Within this platform we use two \textit{interactive datasets}, which allow autonomous robot navigation, built from 3D scans of real-world environments: Matterport3D (MP3D) \cite{Matterport3D}, and Habitat-Matterport 3D Research Dataset (HM3D) \cite{ramakrishnan2021hm3d}. Each of these datasets consist of 30.22k and 112.50k $m^2$ of overall navigable space, respectively. MP3D provides 90 different building-scale scenes. HM3D is the much larger with 1,000 building-scale residential, commercial, and civic spaces. In our experiments we use the default train/validation/test split sets provided by the Habitat-Sim setup. This results in 61/11/18 scenes for MP3D, and 800/10/100 for HM3D.

\begin{figure}[!t]
    \includegraphics[width=\columnwidth]{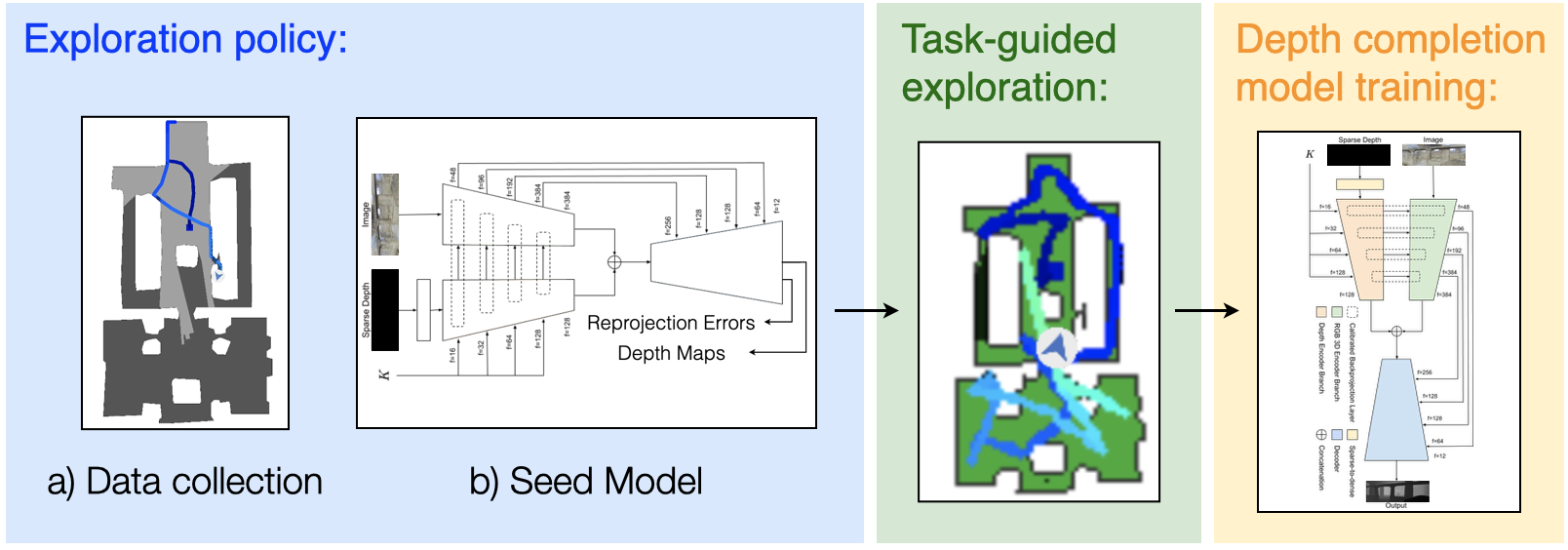}
    \caption{\textbf{Overview of our approach}. Existing work on navigation trains an RL-based Policy to guide exploration \cite{ramakrishnan2021exploration}, we instead train an Seed Model that yields residual poses ($p_{\delta}$) for our navigation policy (Alg. 1).}
    \label{approach}
\end{figure}

\textbf{Data Collection Pipeline}. We follow standard practice for data collection \cite{savva2019habitat,ramakrishnan2021exploration} and deploy a robot, equipped with a navigation algorithm (Fig. \ref{approach}-a), for a maximum of 500 time steps per scene. Our data pipeline comprises a four stage process: robot exploration, sparse depth sampling, data verification, and data preprocessing. The action space of the robot consist of four discrete actions: go forward, turn left, turn right, or stop. We instantiate a robotic agent that has access to a continuous sensory input stream of rendered RGB-D frames and 6D poses of its camera. RGB images and ground-truth depth maps are rendered at resolution 400$\times$400 pixels. We highlight that we do not use these ground-truth dense depth maps for training, and obtain instead extremely sparse depth maps by sampling  $\sim$1500 sparse depth points via Harris corner detector \cite{harris1988combined} (covering $\sim$1.0\% of the full depth map). This allow us to mimic sparse depth maps produced by SLAM/VIO systems \cite{wong2021unsupervised}. 


\textbf{Evaluation Metrics}. We address depth completion tasks given a robotic vision scenario, where an embodied agent can explore a particular scene using autonomously navigation algorithms. 
We allow the robot to use several navigation policies to explore a 3D scene and collect data. This is in contrast to existing evaluations where the model is trained and tested using the same exploration policy (\textit{e.g.}, human defined path); here our benchmark evaluates models on data collected from unseen policies. This would allows us to better evaluate the generalization capabilities of depth completion algorithms to novel viewpoints and scenes. For evaluating depth completion performance, we use four standard metrics: MAE, RMSE, iMAE, and iRMSE (we refer to the read to \cite{wong2021unsupervised} where these are defined, if necessary).

\begin{table*}[!t]
    \caption{Depth Completion Quantitative Results on the MP3D test set}
    \begin{tabular}{lrrrrrrrrrrrrrrrr}
    \toprule
    \multirow{1.8}{*}{Exploration} & \multicolumn{4}{c}{KBNet \cite{wong2021unsupervised}} & \multicolumn{4}{c}{VOICED \cite{wong2020unsupervised}} &\multicolumn{4}{c}{FusionNet \cite{wong2021learning}} & \multicolumn{4}{c}{ScaffNet \cite{wong2021learning}} \\
    \cmidrule(rl){2-5} \cmidrule(rl){6-9} \cmidrule(rl){10-13} \cmidrule(rl){14-17}
    Method & MAE & RMSE & iMAE & iRMSE & MAE & RMSE & iMAE & iRMSE & MAE & RMSE & iMAE & iRMSE & MAE & RMSE & iMAE & iRMSE \\
    \midrule
    Random & 176.27 & 350.95 & 145.83 & 319.16 & 371.29 & 529.80 & 272.62 & 430.13 & 286.70 & 475.37 & 471.49 & 1130.54 & 312.73 & 505.68 & 508.15 & 1221.84 \\
    Frontier & 160.01 & 324.22 & 129.43 & 258.71 & 303.67 & 495.62 & 180.09 & \textbf{296.53} & 210.46 & 417.18 & \underline{196.39} & \underline{440.67} & 205.00 & 412.66 & 200.95 & 469.73 \\
    Oracle & 147.31 & 321.93 & 118.70 & 248.23 & 285.49 & 465.33 & \underline{179.28} & \underline{304.58} & \underline{195.43} & \underline{387.91} & 199.44 & 526.89 & \underline{192.14} & \underline{387.41} & 206.34 & 554.56 \\
    \midrule
    Curiosity & 158.26 & 323.83 & 161.89 & 351.04 & 327.17 & 490.96 & 252.41 & 430.02 & 254.47 & 450.51 & 282.15 & 697.10 & 244.77 & 453.48 & 317.82 & 826.63 \\
    Sem. Rec. & 165.67 & 333.25 & 125.46 & 265.13 & 304.11 & 471.92 & 216.59 & 371.78 & 218.73 & 428.66 & 212.50 & 555.67 & 220.30 & 415.13 & \underline{158.32} & \textbf{319.92} \\
    Coverage & 138.92 & 311.10 & 114.55 & 253.35 & \underline{278.83} & \underline{446.30} & 189.69 & 327.82 & 263.48 & 463.57 & 378.86 & 875.99 & 271.44 & 485.30 & 441.82 & 1097.78 \\
    Novelty & 136.52 & \underline{297.49} & 115.04 & 254.25 & 309.27 & 487.83 & 190.78 & 311.33 & 242.34 & 438.50 & 248.30 & 581.43 & 225.32 & 425.08 & 266.31 & 737.49 \\
    Sm. Cov. & \underline{135.85} & 307.22 & \underline{102.82} & \underline{228.65} & 285.73 & 457.69 & 182.91 & 314.34 & 213.93 & 416.16 & 376.81 & 966.07 & 205.69 & 404.00 & 194.91 & 562.61 \\
    \midrule
    {\tt \textbf{DEUX}} & \textbf{121.07} & \textbf{269.18} & \textbf{84.01} & \textbf{179.10} & \textbf{244.74} & \textbf{415.07} & \textbf{175.05} & 317.55 & \textbf{191.55} & \textbf{386.97} & \textbf{145.19} & \textbf{314.56} & \textbf{176.21} & \textbf{369.90} & \textbf{140.18} & \underline{320.48} \\
    \bottomrule
    \end{tabular}
    \label{table_mp3d}
    \footnotesize{\textbf{bold}: best; \underline{underline}: second best; Sm. Cov.: Smooth Coverage; Sem. Rec.: Semantic Reconstruction}\\
\end{table*}

\textbf{Conventional Exploration}. We use conventional \textit{classic} and \textit{learned} navigation paradigms, typically designed for ensuring full area coverage or point-goal localization tasks, for data collection and benchmarking. We use 8 different navigation algorithms: 3 heuristic-based (Random, Frontier, and Oracle) \cite{yamauchi1997frontier, ramakrishnan2021exploration} and 5 reinforcement learning-based (Curiosity, Semantic Reconstruction, Coverage, Smooth Coverage, and Novelty) methods \cite{ramakrishnan2021exploration}. The Random agent is the simplest baseline that uniformly samples random navigation actions. This agent was found to get stuck when the robot bumped into certain obstacles/objects within a scene, thus we added some heuristics to unstuck the agent when this happens. The Frontier agent is based on the classic frontier-based exploration \cite{yamauchi1997frontier,ramakrishnan2021exploration} that uses the A-Star planning algorithm \cite{hart1968formal} to visit the frontiers (\textit{i.e}, edges between unexplored and free spaces). The Oracle agent is designed to provide an upped-bound exploration performance by having access pre-sampled, specific target locations to navigate \cite{ramakrishnan2021exploration}.

The remaining five learned exploration agents are based on RL training. We use the pretrained RL navigation policies provided by \cite{ramakrishnan2021exploration}, where each policy is trained with a specific reward functions using the PPO algorithm \cite{schulman2017proximal}. A brief description of each policy is provided as follows. Curiosity: encourages visiting poorly predicted states by a forward-dynamics model. Object-based semantic reconstruction: rewards visiting states that allow better semantic reconstruction. Coverage: maximizes the overall area visited. Novelty: rewards visiting less frequently visited states. Smooth Coverage: bridges the gap between Novelty and Coverage by rewarding the number of times a region was observed. This allows the discovery of unexplored parts by navigating across less frequently visited locations. We refer the reader to \cite{ramakrishnan2021exploration} for additional details.

\textbf{Proposed Exploration Approach}. In addition to these conventional navigation paradigms, we collect training data with our proposed active depth-guided exploration approach. Similar to conventional learning-based paradigms, which requires training an \textit{RL policy}* (Fig. \ref{approach}-b), our depth-guided navigation approach trains a depth \textit{seed model} (Fig. \ref{approach}-b) which enables visiting locations with high depth uncertainty using our navigation policy (Alg. 1). Thus, we first use a Random agent to collect data and train our seed model (Fig. \ref{approach}-a,b). We then train our depth completion model (Fig. \ref{approach}-right) from scratch using only the data collected in our depth-guided exploration stage (Fig. \ref{approach}-middle). We highlight that although both the seed model and the depth completion model share the same architecture, they have different functionalities and are not used interchangeably. We also note that the seed model can be replaced by any other pretrained depth model that can be taken off-the-shelf, \textit{e.g.}, not necessarily requiring weights trained on a random agent. Furthermore, our approach is agnostic to the depth completion model chosen, supervised or unsupervised, as we report in Section \ref{sec:results} and Tables \ref{table_mp3d}, \ref{hm3d}.

\textbf{Depth Completion Training Details}. We evaluate the depth completion performance yield by conventional navigation methods and our proposed depth-guided exploration approach using three unsupervised (KBNet \cite{wong2021unsupervised}, VOICED \cite{wong2020unsupervised}, FusionNet \cite{wong2021learning}) and one supervised (ScaffNet \cite{wong2021learning}) depth completion models. The predicted and evaluated depth values were both set within the range of [0.1, 10.0] meters. We train these models, using 384$\times$384 crops, for 25 epochs with the Adam optimizer \cite{kingma2014adam} and a learning rate of $1e^{-4}$ for the initial 10 epochs and then $5e^{-5}$; as we found through cross validation experiments that those worked well for each particular model. The remaining hyperparameters are set following their respective papers. For KBNet, VOICED and FusionNet, training on MP3D takes $\sim$2h and on HM3D $\sim$24h, while for ScaffNet $\sim$0.5h and $\sim$6h, respectively. This is using a single 20GB NVIDIA RTX A4500 GPU with a 24-Core AMD Ryzen 3960X CPU. 

\subsection{Results} \label{sec:results}

We compare the performance of depth completion  models trained on data collected by several types of navigation paradigms, along with our proposed depth-guided exploration approach. Table \ref{table_mp3d} shows the results of \textit{unsupervised} (KBNet, VOICED, FusionNet) and \textit{supervised} (ScaffNet) methods on the MP3D test set.

\textbf{Classic Exploration}. In Table \ref{table_mp3d}, for the KBNet model for instance, the Frontier and Oracle methods outperform the Random agent. This is expected, as these agents are designed to visit specific locations, and thus are able to collect more informative data. The Oracle agent, which has access to specific goal references per scene (pre-set by a human expert), is significantly better (in all metrics) than the Frontier agent, which only performs heuristic-based search; as described in the previous Subsection. This similar trend is also observed for VOICED, FusionNet and ScaffNet, although not across all metrics as KBNet is a recent work.

\textbf{Learned Exploration}. The use of learning-based exploration policies substantially improves the depth completion performance of classic agents. For instance, in Table \ref{table_mp3d} for KBNet, we note that the Smooth Coverage agent outperforms the Oracle and Random agents by 7\% and 22\% in terms of MAE, respectively. Now, the fact that the Smooth Coverage (Sm. Cov.) approach improves over the Novelty and Coverage agents is not surprising, as it combines the best of both policies in terms of reward functions. As per the Semantic Reconstruction (Sem. Rec.) and Curiosity agents, their lower performance, among the learned agents, is likely due to the inherently sparser nature of its reward functions, as these are defined by predicted semantic-based object locations and novel predicted robot states, respectively.

\textbf{Proposed Depth-guided Exploration}. With the insights provided by benchmarking existing navigation algorithms. We now evaluate the performance of our proposed depth-guided exploration agent. As shown in Table \ref{table_mp3d}, our proposed agent outperforms all the exploration methods by significant margins using either unsupervised or supervised depth completion models. Overall, our exploration approach outperforms by 19\%, 18\% and 48\% to existing learning-based (semantic reconstruction), heuristics (oracle), and random exploration techniques, respectively, across all metrics and depth models. This demonstrates that our agent is visiting locations with high depth uncertainty and collecting more informative data specifically for depth perception tasks.

\begin{figure}[!t]
    \centering
    \includegraphics[width=\columnwidth]{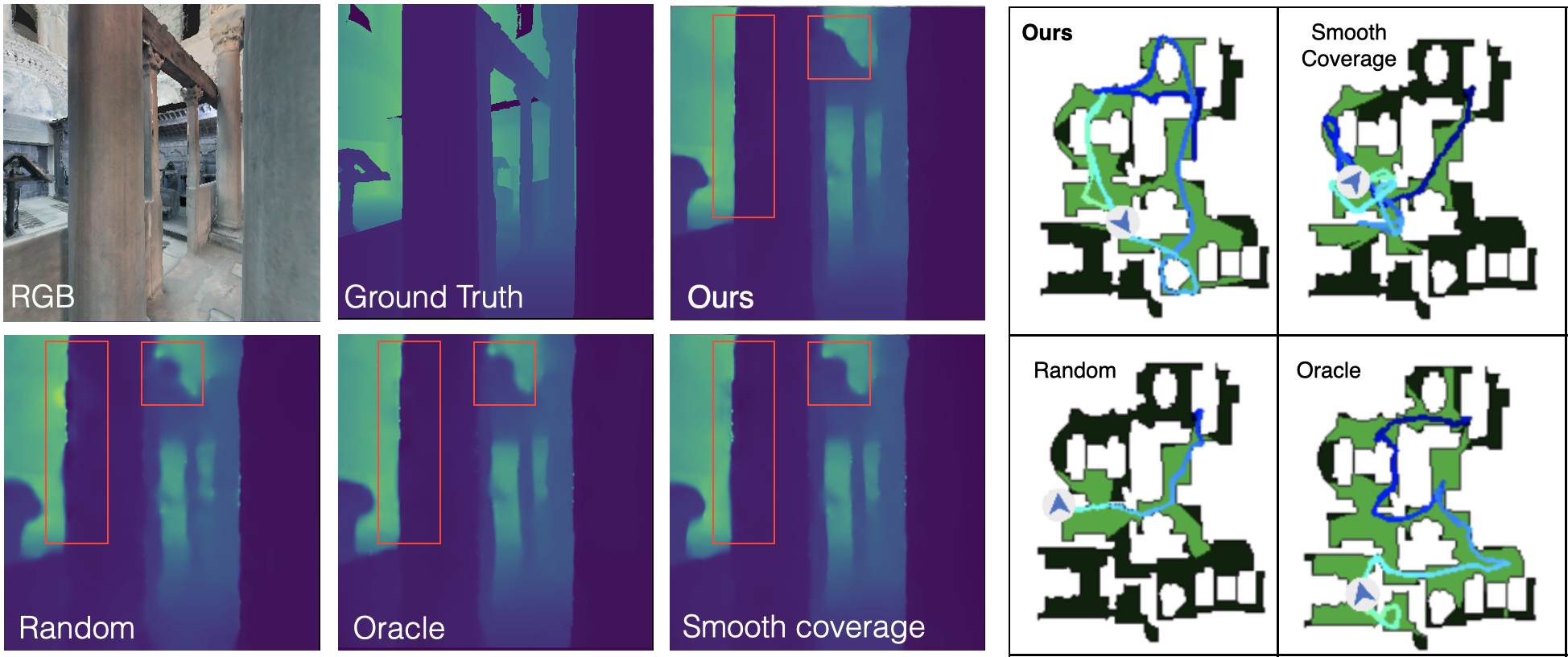}
    \caption{\textbf{Left}: Qualitative depth completion results of KBNet (sharper depth maps: better performance) using heuristic exploration (Random, Oracle), learning-based navigation (Smooth Coverage), and our approach (DEUX) on the MP3D test set. \textbf{Right}: Visualization of exploration policies: Top-down view navigation trajectories on the MP3D test set (black: unexplored, green: explored). Color changes in the trajectories represent time progression.}
    \label{mp3d_results}
\end{figure}

Fig. \ref{mp3d_results}-left shows qualitative depth completion results, from Table \ref{table_mp3d}, for KBNet. And Fig. \ref{mp3d_results}-right shows the qualitative robot exploration trajectories obtained for their respective navigation paradigm during data acquisition, also for KBNet. Here we highlight that existing exploration methods such as smooth coverage (best of learning-based methods) or oracle (best of heuristic techniques), which are particularly designed to maximize specific reward functions or cover larger areas, respectively, do not necessarily achieve good depth estimation. In contrast, our task-driven exploration approach shows that visiting locations with high depth uncertainty yields significantly better depth completion results.

Table \ref{hm3d} reports the results on HM3D. We note that for this particular dataset we do not compare against the remaining navigation paradigms (as in Table \ref{table_mp3d}) given that those exploration agents are not provided/trained on HM3D in \cite{ramakrishnan2021exploration}. However, in Table \ref{table_mp3d_on_hm3d}, we provide zero-shot generalization results of all the exploration methods (obtained by training all depth models on MP3D; from Table \ref{table_mp3d}). Additionally, Table \ref{hm3d} reports the zero-shot generalization results (\textit{i.e}., depth models trained on MP3D with {\tt DEUX}*), which yields better results compared to a random exploration approach trained on HM3D. Across all metrics, we observe similar trends for our approach ({\tt DEUX}) than those obtained in Table \ref{mp3d_results}, outperforming a heuristic-based paradigm by over 40\% on average across all metrics and depth models.

\textbf{Zero-Shot Generalization}. In Table \ref{table_mp3d_on_hm3d} we show cross-dataset generalization results, \textit{i.e.}, train on MP3D and evaluate on HM3D. We evaluate each depth model, trained on MP3D from Table \ref{table_mp3d}, on the HM3D test set, and report the average for all models (KBNet, VOICED, FusionNet, and ScaffNet) across each metric. Overall, our approach outperforms all existing exploration methods including smooth coverage (best of RL-based exploration), oracle and random agents by 5\%, 17\% and 24\%, respectively. Note that these results are for zero-shot transfer learning alone.

\begin{table}[!t]
    \caption{Depth Completion Quantitative Results on the HM3D test set}
    \begin{tabular}{llcccc}
    \toprule
    Depth Model & Exploration Method & MAE & RMSE & iMAE & iRMSE \\
    \midrule
    \multirow{3}{*}{KBNet \cite{wong2021unsupervised}} & Random & 80.33 & 170.12 & \underline{394.45} & 757.24 \\
    & {\tt DEUX}* &  \underline{71.40} & \underline{138.04} & 442.31 & \underline{713.86} \\
    & {\tt \textbf{DEUX}} &  \textbf{57.71} & \textbf{111.22} & \textbf{246.10} & \textbf{440.81} \\
    \midrule
    \multirow{2}{*}{VOICED \cite{wong2020unsupervised}} & Random & 143.64 & 226.33 & \underline{611.88} & \underline{846.21} \\
     & {\tt DEUX}* & \underline{131.01} & \underline{213.86} & \textbf{510.25} & \textbf{714.48} \\
    & {\tt \textbf{DEUX}} & \textbf{119.52} & \textbf{202.14} & 709.58 & 961.92 \\
    \midrule 
    \multirow{2}{*}{FusionNet \cite{wong2021learning}} & Random & 133.91 & 232.82 & 967.09 & 1690.78 \\
    & {\tt DEUX}* & \underline{103.90} & \underline{191.07} & \underline{763.54} & \underline{1091.26} \\
    & {\tt \textbf{DEUX}} & \textbf{80.60} & \textbf{157.72} & \textbf{371.78} & \textbf{726.19} \\
    \midrule
    \multirow{2}{*}{ScaffNet \cite{wong2021learning}} & Random & 136.51 & 236.98 & 1045.77 & 1851.53 \\
    & {\tt DEUX}* & \underline{102.29} & \underline{196.78} & \underline{592.42} & \underline{918.72} \\
    & {\tt \textbf{DEUX}} & \textbf{84.82} & \textbf{163.84} & \textbf{391.78} & \textbf{811.71} \\
    \bottomrule
    \end{tabular}
    \label{hm3d}
    \footnotesize{*Trained on MP3D. \textbf{bold}: best; \underline{underline}: second best}\\
\end{table}

\begin{table}[!t]
    \caption{Zero-Shot Transfer (cross-dataset generalization)}
    \centering
    \begin{tabular}{lrrrrrrrrrrrrrrrr}
    \toprule
    Exploration Method & MAE & RMSE & iMAE & iRMSE \\
    \midrule
    Random & 124.43 &	206.05 &	652.46 &	1090.34 \\
    Frontier & 126.27 &	212.21 &	865.61 &	1218.59 \\
    Oracle & 116.93 &	202.37 &	769.39 &	1157.59 \\
    \midrule
    Curiosity & 113.48 &	192.76 &	\underline{577.13} &	\underline{906.99} \\
    Semantic Reconstruction & 114.79 &	197.41 &	634.19 &	987.14 \\
    Coverage & 113.66 &	196.22 &	597.73 &	990.14 \\
    Novelty & 108.85 &	190.04 &	602.79 &	956.48 \\
    Smooth Coverage & \underline{105.52} &	\underline{187.92} &	649.50 &	1052.36 \\
    \midrule
    {\tt \textbf{DEUX}} & \textbf{102.15}	& \textbf{184.93} &	\textbf{555.61} &	\textbf{859.58} \\
    \bottomrule
    \end{tabular}\\
    \label{table_mp3d_on_hm3d}
    \footnotesize{\hspace{-32mm}\textbf{bold}: best; \underline{underline}: second best}\\
\end{table}

\textbf{Influence of Exploration Paradigms}. Now we compare the influence of existing paradigms for handling the exploration and data collection phases, see Fig. \ref{approach}, introduced in \cite{ramakrishnan2021exploration} (area coverage, novelty, curiosity, and semantic reconstruction) against our approach. From Table \ref{table_mp3d}, for the best performing model (KBNet), we highlight that our exploratory approach yields better results than conventional paradigms as follows:  {\tt DEUX} $>$ Smooth Coverage $>$ Novelty $>$ Sem. Rec. $>$ Curiosity. This provides a key insight on the utility of navigation paradigms for learning specific downstream tasks such as depth completion. Competent depth completion learning was achieved when the robot/camera visited locations with high depth uncertainty, which are not necessarily those with high novelty or low coverage within an environment.

\section{CONCLUSIONS} \label{sec:conclusions}

We presented a novel exploration paradigm for unsupervised depth perception tasks in the context of robotic navigation. Our proposed task-informed exploratory approach is based on photometric reprojection residuals, which are used for robot path planning across locations with high depth uncertainty. Extensive experimental results using four depth completion models on two interactive datasets have shown that our approach outperforms existing (classic and learning-based) exploration techniques by significant margins. Our key insight is that existing navigation paradigms, typically used for data collection and training of a wide range of robotic vision applications, do not necessarily provide task-specific data points to achieve competent learning for a particular downstream application. In future work includes extending this to other robot perception tasks both in simulated environments as well as real world deployment.

\addtolength{\textheight}{-1.7cm}   

\bibliographystyle{IEEEtran}
\bibliography{references}

\end{document}